# Evaluation of an indoor localization system for a mobile robot

Víctor J. Expósito Jiménez
*Dependable Systems Group*
*Virtual Vehicle Research Center*
Graz, Austria
victor.expositojimenez@v2c2.at

Christian Schwarzl
*Dependable Systems Group*
*Virtual Vehicle Research Center*
Graz, Austria
christian.schwarzl@v2c2.at

Helmut Martin
*Dependable Systems Group*
*Virtual Vehicle Research Center*
Graz, Austria
helmut.martin@v2c2.at

*Abstract*—Although indoor localization has been a wide researched topic, obtained results may not fit the requirements that some domains need. Most approaches are not able to precisely localize a fast moving object even with a complex installation, which makes their implementation in the automated driving domain complicated. In this publication, common technologies were analyzed and a commercial product, called Marvelmind Indoor GPS, was chosen for our use case in which both ultrasound and radio frequency communications are used. The evaluation is given in a first moment on small indoor scenarios with static and moving objects. Further tests were done on wider areas, where the system is integrated within our Robotics Operating System (ROS)-based self-developed "Smart PhysIcal Demonstration and evaluation Robot (SPIDER)" and the results of these outdoor tests are compared with the obtained localization by the installed GPS on the robot. Finally, the next steps to improve the results in further developments are discussed.

*Index Terms*—indoor localization, robotics, ROS, automated driving

## I. Introduction

Indoor localization has been a highlighted topic since more technologies are capable of offering a relatively good precision and can be used in a wide number of applications. In the context of automated driving, outdoor localization has been mainly focused on Global Navigation Satellite System (GNSS) that offers proved results in most scenarios [1] [2]. Unfortunately, indoor localization is one of the scenarios in which GNSS cannot be used because the signal strength is far from being reliable inside a building. Thus, several approaches have been researched along these years to find a solution to this problem.

On the other side, nowadays the implementation of robots is not only limited to the industry such as the typical arm-robot, which supports humans in heavy loads or repetitive and precise tasks. With the development of many open-source developments, a robot can be used on almost any scenario due to the availability of developments such as Robotics Operating System (ROS) [3]. ROS is an open-source middleware that allows to build robot applications using a wide range of third-party libraries and tools for different domains. For example, Autoware.auto [4] which adapts ROS to be used specifically on autonomous cars.

The main goal of this publication is to give an overview of the existing technologies and the implementation of a reliable and feasible system, which allows to precisely localize mobile robots in indoor environments. In the next section, the motivation and related work are given. Section III describes the robot in which the proposed indoor localization system is implemented to evaluate it in section IV. Finally, Section V describes our conclusion according to the obtained results and the further steps to follow.

## II. Motivation

Many publications have studied the feasibility of different technologies, from most common QR Codes/Aesthetic markers [5] or lasers [6] to the proposed solutions based on radio communications, which is the main scope of this publication. [7] depicts a well-detailed overview of the most commonly used approaches and technologies for indoor localization. A more focused overview on Radio Frequency Identification (RFID) is given in [8]. Low (LF) (125-134 kHz) and High Frequencies (HF) (13.56 MHz) have been rejected since they present a short reading range and only a few tags can be read at the same time. On the other hand, Ultra High Frequency (UHF) signals are weaker in environments with water or metals as well as on the penetration in a human/animal body. The publication describes a detailed overview of the location techniques that can be used with UHF RFID. There are two kind of applications based on this technology, reader localization and tag localization. Reader applications are used when each localized object includes a RFID reader. This method is more expensive once every device needs a reader but, nonetheless, presents a better accuracy. Tag localization is mostly used because passive UHF tags are much cheaper than readers. Unfortunately, the precision of this localization method is worse (one meter in the best of cases) and the moving objects cannot be as fast as in the reader localization.

Bluetooth technology is another well-known technology for indoor localization. The approach detailed in [9] is able to reach a precision less than two meters on vehicle indoor localization by combining Bluetooth Low Energy (BLE) and sensor data from the Controller Area Network (CAN). Another research publication [10] confirms these results regarding Bluetooth. Here, the authors combine Bluetooth beacons with proposed algorithms to improve the accuracy but they are not able to reach a precision of less than a few meters.

TABLE I
RADIO COMMUNICATION LOCALIZATION TECHNOLOGY OVERVIEW

| Technology | Accuracy | Range | Price | Frequency |
|---|---|---|---|---|
| UHF RFID reader localization (passive) | 0.1 – 1m | 10 – 15m | High | 860MHz – 960MHz |
| UHF RFID tag localization (passive) | 1 – 5m | 10 – 15m | Low | 860MHz – 960MHz |
| Bluetooth | 2 – 3m | 10 – 15m | Low | 2.4GHz |
| ZigBee | 1 – 5m | 10 – 40m | Low | 868MHz – 960MHz, 2.4GHz |
| WLAN | 1 – 5m | 10 – 40m | Medium | 2.4GHz |
| Ultra Wide Band (UWD) | 0.1 – 0.2m | 10 – 50m | High | 3.1GHz – 10.6GHz |

Several approaches are built based on ZigBee [11] [12] [13] or WLAN [14] [15]. These approaches offer relatively cheap solutions with a good range as well as a one to ten meters accuracy. One of the advantages of ZigBee over WLAN is the usage of the Industrial, Scientific, and Medical (ISM) bands, 915MHz (USA), and 868MHz (EU); which avoids the congested public 2.4GHz band. Besides, [16] presents a comparison of these technologies with the usage of Ultra-Wide Band (UWD) to localize objects. Unfortunately, the difficulty to localize a mobile object and its expensive installation represent big obstacles.

Ultrasounds have been also used to scope this problem. In the proposed approach by [17], the authors reach a precision up to five centimeters on stationary objects. Furthermore, a combined approach of ultrasounds and RFID is shown in [18] [19]. Here, authors are able to get an accuracy of about fifty millimeters but it is only tested on small distances around three meters. An interesting combination of technologies is also given in the product Marvelmind Indoor GPS [20] which uses both ultrasound and radio communication through the ISM radio bands to obtain a precise localization of an object within two centimeters accuracy.

Table I shows a comparison of the different technologies presented in this work, which summarizes what these technologies are capable of. The chosen indoor localization system for our use case has to fulfill most of the requirements that are given in the following list:

- Precision/Accuracy: less than 0.1m in static objects, 0.1 – 1m for mobile objects.
- Technology: Radio.
- Range: 20 – 25 meters.
- Maximum speed: 15km/h.

If the table is revisited, technologies that would allow us to reach these requirements were not found. UHF RFID reader localization could be used but the usage of a reader for each object makes the implementation very expensive due to the cost of the devices and the complex installation due to the big and heavier components. UWD could also be a possible but, similar to the RFID reader localization, the cost devices and the complex installation are disadvantages that make this technology not available for us. Other presented technologies just do not reach the precision that our use case needs. For all of this, the indoor localization system by Marvelmind theoretically seems to be the only a plausible solution for us since it can reach an accuracy of two centimeters, and the easy installation in which the system is built with small and lightweight components, and costs for a small system are within our budget.

III. SMART PHYSICAL DEMONSTRATION AND EVALUATION ROBOT (SPIDER)

The proposed indoor localization system is implemented in our *"Smart PhysIcal Demostration and Evaluation Robot" (SPIDER)* as shown in Figure 1 which is a self-developed mobile platform for the development and evaluation of Automated Driving (AD) functions.

SPIDER is designed to be used in a wide variety of scenarios and prepared to work in adverse environmental conditions such as rain or fog. It is also used for testing purposes and offers high flexibility to integrate different types of sensors, allowing e.g. the comparison of the measured data with calculated data obtained from a simulation. Moreover, it has an extensible design in which more modules with different sensors can be mounted in an easy way, providing developers the needed flexibility to test new sensors or scenarios in the future. Its purpose is driving scenarios under different weather conditions, allowing customers to efficiently analyze and compare their impact on sensor combinations and their mounting positions.

Additionally, the SPIDER allows the developers high maneuverability to execute tests with no time losses. This is done because of the usage of four independent wheels which gives SPIDER the possibility of omnidirectional movements, including 360° or sidewalks movements. The safe execution of test drives is critical to avoid human harm. For this reason, the SPIDER provides several safety functions implemented on hardware and software level and, for example, ensures the avoidance of collisions.

SPIDER integrates a High-Performance Computer (HPC) with a dedicated Graphical Process Unit (GPU). It controls the whole system and allows to execute high-demanding process such as AI algorithms. The Low-Level Control (LLC) is carried out by the Automotive Realtime Integrated NeXt Generation Architecture (AURIX) [21] platform. ROS is used for the High Level Control (HLC). The usage of ROS gives great flexibility because it is able to use the extensive catalog of third-party libraries that can be used for different purposes and hardware. The PC uses Ubuntu 16.04 as an operating system and the Kinetic version of ROS.

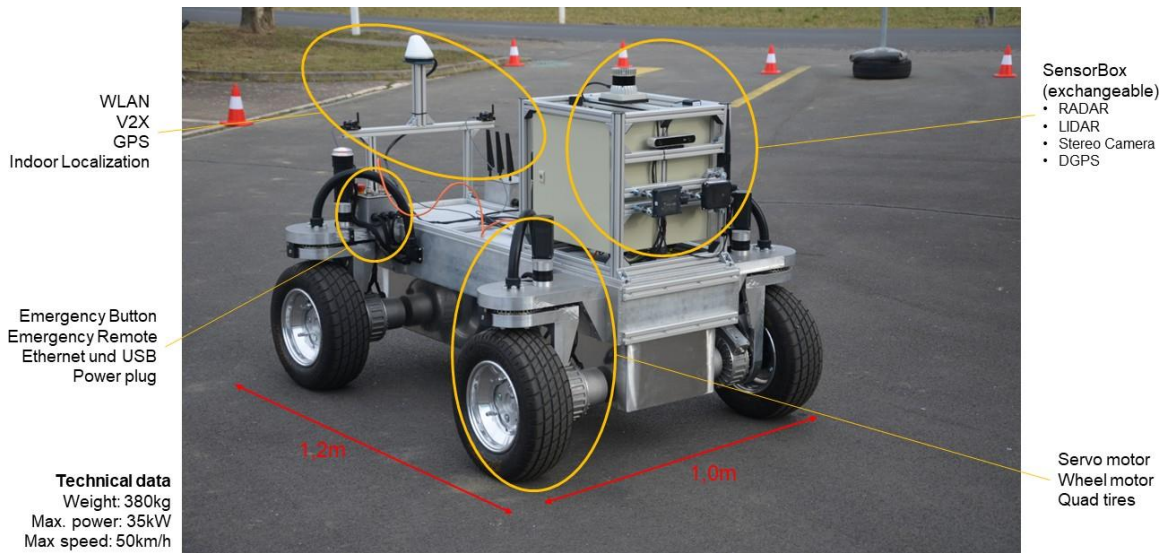

Fig. 1. Smart PhysIcal Demonstration and Evaluation Robot (SPIDER)

In this context, the proposed indoor localization system is integrated in the SPIDER and will be evaluated and compared with the other sensors already installed. In the end, the main goal is the fusion of all installed sensors from the SPIDER, allowing it to autonomously reach a destination point without any human interaction.

## IV. EVALUATION

This indoor localization method uses three main components, namely a modem, stationary beacons and mobile beacons.

- Modem: It is used to set up and monitor the system. The communication between beacons and the modem is done through radio communication, in our case, on the 433 Mhz band.
- Stationary beacons: They act as receivers of the mobile beacons and their locations are previously set up in the modem map. The communication between beacons are done through ultrasounds.
- Mobile beacons: They are the localized objects and transmit their position to the stationary beacons.

Unfortunately, the communication and the developed algorithms to calculate the distance are not open-source and their analysis and description could not be included in this publication. The first tests of this indoor localization system were carried out on two different scenarios: an office and a garage. The office is shown in figure 2 in which the green points depict the situation of the stationary beacons. The configuration for this use case is a submap which includes the four stationary beacons following the recommended configuration by provider. The total covered surface is around a hundred square meters and the table in the picture depicts the distance between each beacon. For this first test, the hedge (mobile beacon) was measured without movement in which it has been measured in different points along the scenario.

According to the obtained results, the mean error in this use case is ten centimeters and the median error was also seven centimeters. On the other hand, other tests were done in a bigger place, a garage. In this case, the total covered surface is a hundred and thirty square meters and, as in previous tests, one submap was configured where four stationary beacons were set up. Similar to the other scenario, the hedge was localized without movement in different points of this garage to know the precision of these measurements. Both mean and median errors were in this case six centimeters. In these first tests, the system presents a result in both scenarios that fits our initial requirements. It can be observed that errors in the garage scenario are smaller than the given in the office although the covered surface of the garage is bigger than the one covered by the beacons on the office room. The reason for this difference could be that the absence of obstacles in the garage favors the measurements since ultrasound needs a clear line of vision to work.

Next step was testing the behavior of the system when mobile objects are localized. The garage scenario with the same stationary beacons configuration was the place for these tests. Figure 3 shows a general behavior of the obtained results. In this use case, the hedge follows a path drawn on the floor, which is painted as an orange line in the figure. This route was done several times in two different velocities, slow (approx. 3km/h) and fast (approx. 8km/h). The system is able to precisely locate the object most of the time when the hedge moves slowly. There is only a position error, around forty centimeters, between beacon number one and four that occurs every time in that position when the hedge moves slowly. If the movement is faster, there is a deviation in that point but not as big as the same scenario presented in slow movement. In this use case, the localization is worse especially when the object turns around due to the lack of measurements at that velocity to successfully calculate the current object path with many direction changes.

Finally, the system was also evaluated in an outdoor scenario. The SPIDER robot could be used, for example, at production facilities in which the changes between indoor scenarios and outdoor scenarios are common, and the behavior has to be proved in this kind of scenario to ensure that the robot localization system is able to successfully handle these changes. This also gives us the possibility to test this indoor localization system on a larger surface than usually available in indoor spaces. Moreover, the results obtained from the proposed localization system can be compared once the GPS positioning is available in outdoor scenarios.

Although four beacons have been set as stationary beacons in previous setups, in this scenario, three beacons were set as stationary beacons due to lack of more available hardware, because two beacons were set as mobile beacons or hedges, which allows us to get not only the position but also the direction of the robot. Two 2D submaps were configured in the modem, one of beacons 1 and 3 and the other one for beacons 1 and 2. The stationary beacons are located in line with a ten meters distance between each one. The used GNSS is a NEO-M8P-2 [22] Module. To improve the satellite localization a base station was also configured in this setup. The base station remains fixed in the scenario and it sends correction data to the rover module which is installed on the robot. The correction data technology is called Real Time Kinematics (RTK) and is able to provide level of precision of ten centimeters when it works on "fixed mode". Unfortunately, we were only able to work on "float mode" in our tests which provides a forty centimeters precision. Figure 4 shows the setup of this scenario in which the location of the stationary beacons, base station, and the total covered area are depicted.

Figure 5 shows the obtained path of the test drives when the maximum reached velocity by the SPIDER is not more than five kilometers per hour. The red path shows the path given by the proposed localization system and the white point the coordinates measured by the rover. According to these tests, the proposed solution is able to locate the robot compared to the GNSS module in most of the path but there are some moments in which the errors become not acceptable for the context of automated driving (approx. one and a half meters). In this case, there are some external factors that could have increased the obtained error of the measures such as the wind bursts during the test influencing the ultrasonic communication between the beacons. Moreover, it has been noticed that at some points, there are not enough measurements by the localization system, especially in the areas farther away from the stationary beacons. This lack of measurements suggests that a more conservative configuration would be needed in which more stationary beacons will be placed to ensure the object is located without any problem along the whole area.

In this first evaluation and according to our results, this localization method fulfills our first expectations, but results have to be improved to reach our requirements since sometimes the localization becomes unstable and the frequency rate is not enough to safely localize the robot. Therefore, the next steps of the research will be the improvement of the obtained results by tuning the configuration parameters and improving the scenario definition.

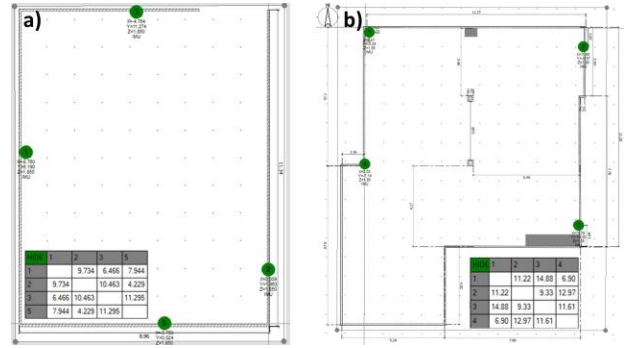

Fig. 2. Indoor test scenarios. a) Office: $100 m^2$ total covered area b) Garage: $130 m^2$ total covered area

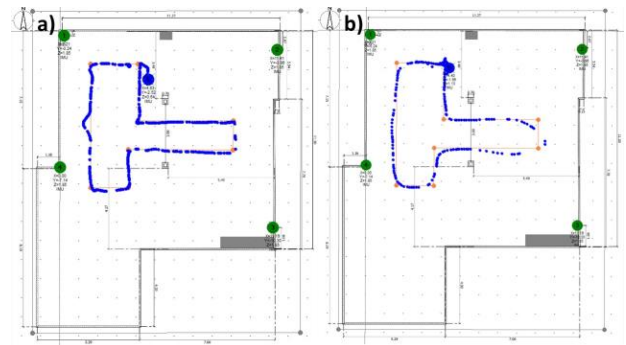

Fig. 3. Obtained path by the indoor localization system with mobile objects. a) Object speed: approx. 3Km/h, b) Object speed: approx. 8Km/h

## V. CONCLUSIONS

In this publication, an overview of the different methods for indoor localization was described and a system, which combines an ultrasound and radio frequency communications, called Marvelmind Indoor GPS was discussed in detail. Next step was the testing of this system to ensure that the system is able to work as expected. First tests were done in two indoor and one outdoor scenarios. The median error in indoor scenarios for static objects were usually less than ten centimeters. In addition to these results, the errors for mobile objects were less than forty centimeters in most of the cases.

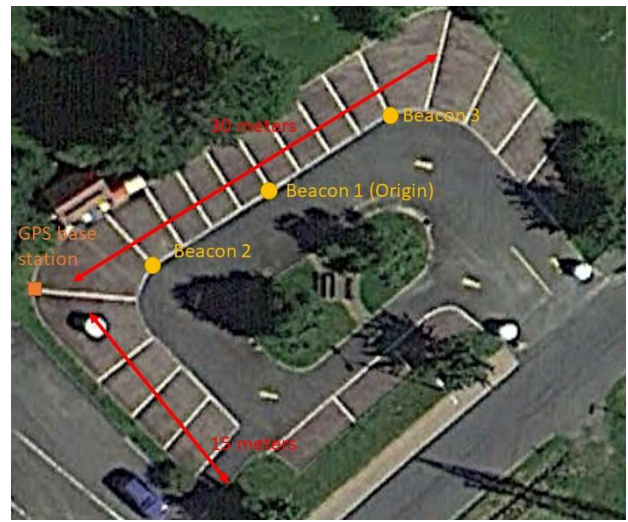

Fig. 4. Outdoor scenario setup

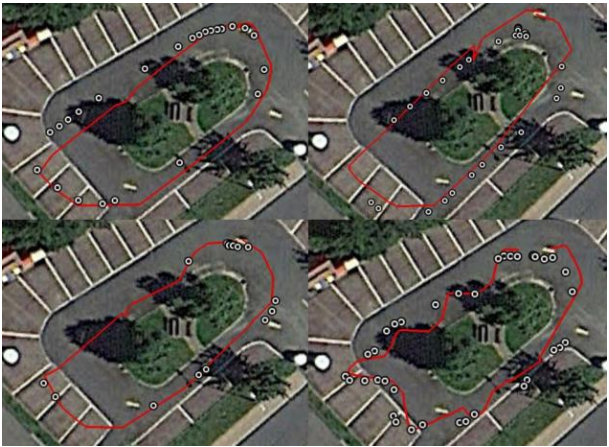

Fig. 5. Results on the outdoor scenario. Points: GPS localization. Red Path: Marvelmind localization.

Outdoor tests were also done where the obtained results were compared to the included GPS from the robot. In this situation in which the scenario is bigger than the used for indoor tests, the comparison between the GPS localization and the proposed solution was similar and the maximum localization differences were between one and one and a half meter.

This paper describes the work-in-progress and is the beginning of the indoor localization implementation for the SPIDER mobile robot in which more tests in more challenged scenarios will be done as well as a better data collection will be implemented to be able to make a better analysis of the obtained data. The localization data of these sensors will be fused with more implemented sensors in the robot to finally reach the robot localization that satisfies our requirements.

ACKNOWLEDGMENT

Research leading to these results has received funding from the EU ECSEL Joint Undertaking under grant agreement n° 737459 (project *Productive4.0*) and from the partners national funding authorities FFG on behalf of the Federal Ministry for Transport, Innovation and Technology (bmvit) and the Federal Ministry of Science, Research and Economy (BMWFW). This research was also partially funded by the Austrian research funding association (FFG) and COMET FFG within the research project *"Autonomous Car To Infrastructure communication mastering adVerse Environments"*(ACTIVE).